\documentclass{article}
\usepackage[final]{colm2026_conference}

\usepackage{microtype}
\usepackage{hyperref}
\usepackage{url}
\usepackage{booktabs}
\usepackage{graphicx}
\usepackage{multirow}
\usepackage{amsmath,amssymb,amsfonts}
\usepackage{amsthm}
\usepackage{mathrsfs}
\usepackage{bm}
\usepackage{xcolor}
\usepackage{algorithm}
\usepackage{algorithmicx}
\usepackage{algpseudocode}
\usepackage{marvosym}
\usepackage[normalem]{ulem}

\definecolor{darkblue}{rgb}{0, 0, 0.5}
\hypersetup{colorlinks=true, citecolor=darkblue, linkcolor=darkblue, urlcolor=darkblue}

\newif\ifincludesupplement
\includesupplementtrue

\let\cite\citep

\theoremstyle{definition}

\theoremstyle{remark}

\title{Proprioceptive--visual correspondence enables\\self-other distinction in humanoid robots}

\author{Yurun Chen$^{1,2}$, \ Tianyuan Gao$^{3}$, \ Yizhong Ge$^{1}$, \ Shikun Ban$^{4}$ \\[0.4em]
\bfseries Yizhou Wang$^{3}$, \ Hongkai Xiong$^{2,5\,\text{\Letter}}$, \ Wenjun Zeng$^{1,6\,\text{\Letter}}$, \ Wentao Zhu$^{1,6\,\text{\Letter}}$ \\[0.6em]
\normalfont $^{1}$Eastern Institute of Technology, Ningbo \quad $^{2}$Shanghai Jiao Tong University \\
$^{3}$Peking University \quad $^{4}$Carnegie Mellon University \\
$^{5}$East China Normal University \quad $^{6}$Ningbo Institute of Digital Twin \\[0.3em]
\small $^{\text{\Letter}}$Corresponding authors}

\begin{document}

\maketitle

\begin{abstract}

Distinguishing self from others is a prerequisite for social intelligence, yet humanoid robots that increasingly share workspaces with humans still lack this ability. Here we show that a humanoid robot can learn self-other distinction from proprioceptive--visual correspondence, without any identity labels or kinematic models. Once established, this distinction bootstraps a predictive self-model that maps joint configurations to three-dimensional body occupancy, capturing how the robot's body changes with action. In multi-agent scenes involving humans or morphologically identical robots, the system reliably identifies itself, learns a 3D self-model, and supports downstream tasks including target reaching, collision-aware motion planning, and human-to-robot motion retargeting. Together, these results outline a route toward bodily self-representation in robots that act and coordinate alongside others in shared physical environments. Project page: \url{https://euron-zc.github.io/humanoid-self-model/}.

\end{abstract}

\section{Introduction}\label{sec:introduction}

\begin{figure*}[t]
    \centering
    \includegraphics[width=0.98\linewidth]{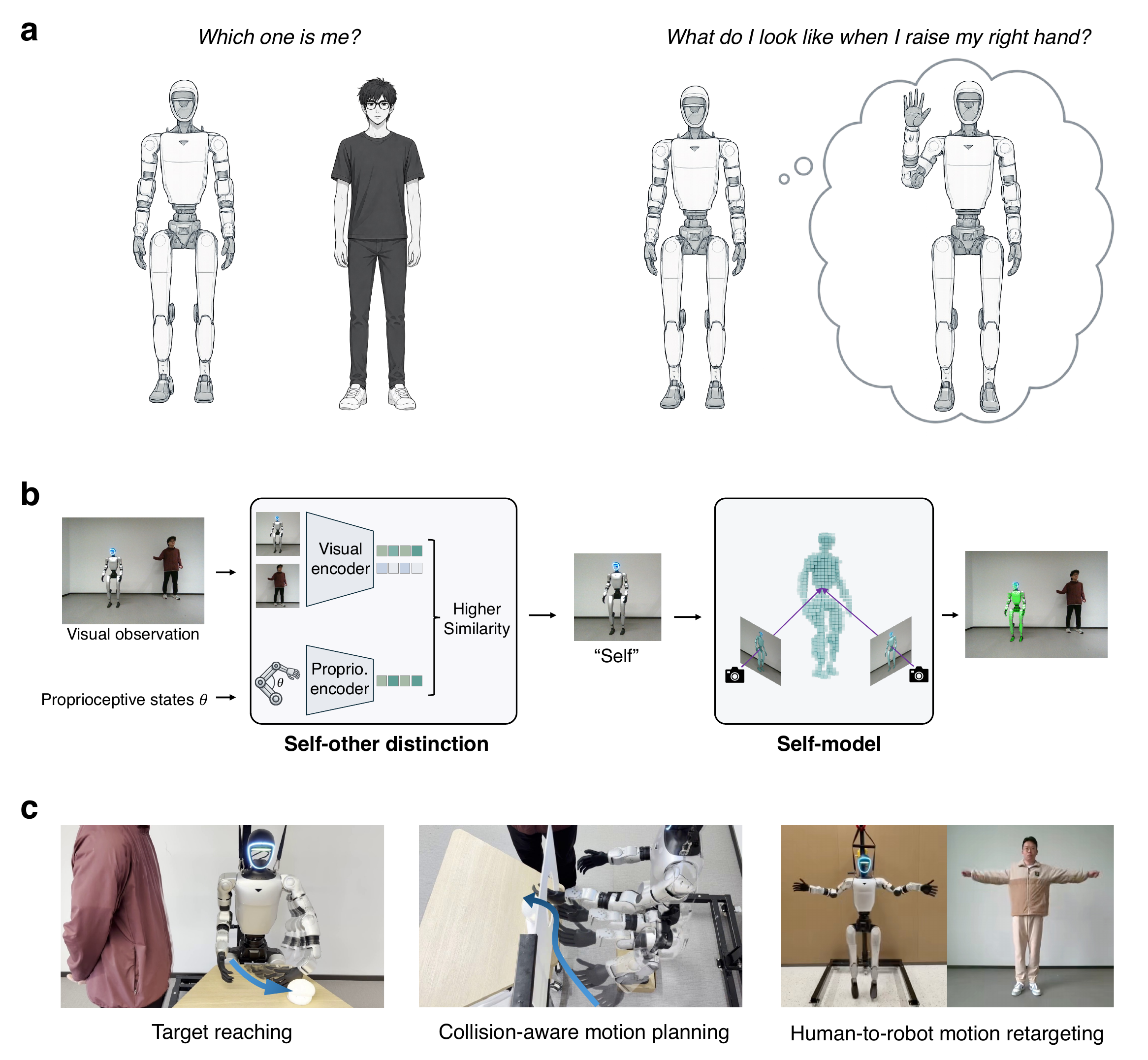}
    \caption{\textbf{Framework overview.} \textbf{a,} A humanoid robot in a shared scene faces two coupled problems: identifying which visible body is its own, and predicting how that body should appear under a given proprioceptive state. \textbf{b,} From only proprioceptive states and visual observations, the proposed framework first distinguishes self from other, then learns a kinematics-free self-model. \textbf{c,} Together, self-other distinction and the learned self-model enable downstream tasks including target reaching, collision-aware motion planning, and human-to-robot motion retargeting.}
    \label{fig:teaser}
\end{figure*}

Distinguishing self from others is a fundamental capability 
for social intelligence~\cite{gallagher2000philosophical,decety2003,steinbeis2016role}. Before an agent 
can imitate a demonstrator, coordinate with a partner, or 
simply avoid colliding with a bystander, it must resolve a 
prior question: ``which body is mine?''
Cognitive science points to proprioceptive--visual 
correspondence as a key mechanism: infants as young as 
three to five months detect the contingency between 
proprioceptive signals and visual feedback of their own 
limb movements, both temporally~\cite{bahrick1985,rochat2000perceived} and 
spatially~\cite{rochat1995, schmuckler1996visual}, and use this match to 
distinguish self from non-self. In adults, synchrony 
between felt and seen body movements can induce body 
ownership even in the absence of tactile 
cues~\cite{kalckert2012}, and subjects distinguish 
self-generated from observed movements by detecting 
mismatches between motor commands and visual 
feedback~\cite{jeannerod2003}. In each case, internal 
bodily signals serve as a private reference against which 
visible bodies are evaluated.

The same question now confronts robots~\cite{gold2009using,stoytchev2011self,lanillos2020robot}. Humanoid robots are entering social environments~\cite{SONG2022489,tong2024advancements,2026Gu}, where they coexist with humans and morphologically identical peers. To act effectively in such settings, they need to solve two coupled problems: \emph{self-other distinction}, identifying which body in the scene is itself, and \emph{self-modeling}, acquiring a predictive representation of that body and how it changes with action~\cite{Bongard2006,Kwiatkowski2019}, a capacity that recent work has shown can be learned from proprioceptive--visual experience in single-agent settings~\cite{Chen2022,schulze2024,hu2025teaching,egohu2025}. 
These two problems are logically coupled: distinguishing self from other presupposes some notion of one's own body, yet learning about one's body presupposes knowing which body to attend to.
In this work, we consider the most natural and the most challenging instance of this problem: an agent that has \emph{no identity label} indicating which body in the scene is its own and \emph{no prior knowledge} of its own morphology: no kinematic model, no geometric description, nothing that specifies what it should look like. From only raw proprioceptive signals and visual observations, it needs to address both questions.

A clue to this chicken-and-egg dilemma comes from developmental psychology: in human children, perceptual self-other distinction emerges before the maturation of detailed body representations~\cite{brownell2007}. Motivated by this ordering, we follow the same progression: we present a framework in which a humanoid robot first visually distinguishes itself from others in multi-agent scenes and then, from that distinction, learns a predictive self-model (Fig.~\ref{fig:teaser}a).

For self-other distinction, our key insight is that the current proprioceptive state should match a specific body in the current frame (though which one is unknown), but is unlikely to match any body observed at a different moment. This structure provides natural contrastive supervision without any explicit identity labels, and we show that it contains sufficient information for learning instance-level discrimination. For self-modeling, the distinguished self-masks supervise a neural occupancy field conditioned on proprioceptive states. The field predicts the probability of arbitrary 3D points belonging to the robot's body, requiring no predefined kinematic model. In this way, self-other distinction provides the learning signal from which the self-model emerges. We show that accurate self-other distinction is sufficient to bootstrap a faithful self-model in a fully self-supervised manner.

We evaluate the framework on a 29-degree-of-freedom (DoF) humanoid robot in both simulated and real-world multi-agent scenes. From only proprioceptive states and visual observations, and without any predefined kinematic model or identity label, the system achieves robust self-other distinction, learns a self-model that predicts 3D body occupancy, and supports downstream tasks including target reaching, collision-aware motion planning, and human-to-robot motion retargeting. These results suggest that proprioceptive--visual correspondence alone provides a sufficient basis for self-other distinction and self-modeling, opening a path in which robots acquire knowledge of their own bodies from experience rather than from manual specification.

\section{Results}\label{sec:results}

We evaluate the framework by asking three questions: whether proprioception and vision alone can distinguish the robot's own body from others without identity labels or kinematic priors; whether the resulting self-mask can supervise a kinematics-free self-model; and whether the learned self-model can support downstream physical interaction tasks. Experiments are conducted on a 29-DoF Unitree G1 humanoid robot observed from a fixed external RGB camera. We conduct both simulated and real-world experiments, which serve complementary purposes: simulation provides a controlled setting with accurate ground truth (GT) for quantitative evaluation, whereas real-world data test whether the framework remains robust under realistic conditions and human-robot interaction.

\subsection{Distinguishing self from others with proprioception}

\begin{figure*}[t]
    \centering
    \includegraphics[width=0.98\linewidth]{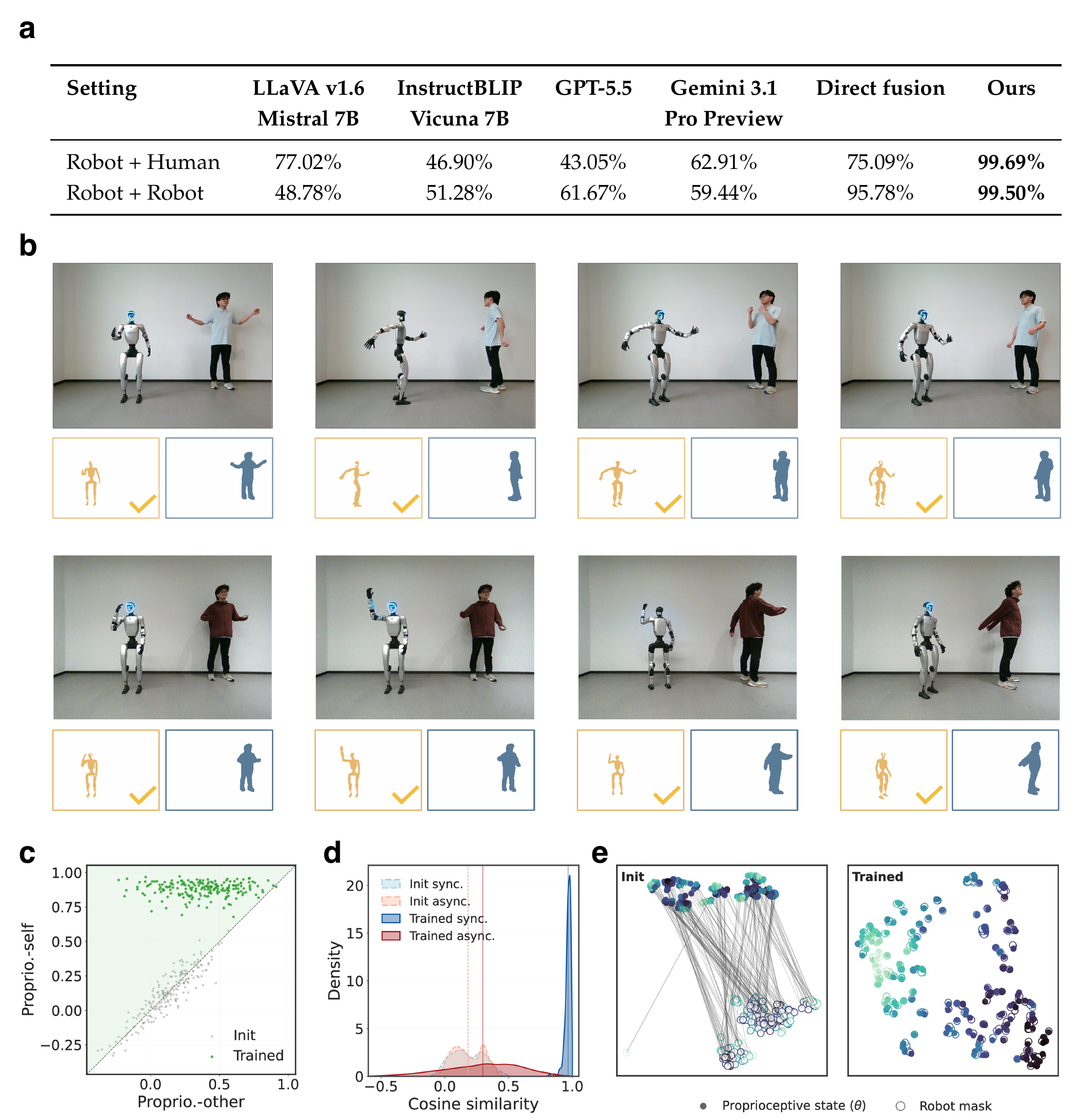}
    \caption{\textbf{Self-other distinction from proprioception.} \textbf{a,} Held-out accuracy across methods in two evaluation settings: a human-robot scene and a two-humanoid scene with a morphologically matched distractor. Comparisons include direct fusion training and vision-language baselines. \textbf{b,} Qualitative examples in which the robot selects its own candidate mask from a scene containing a distractor; the selected mask is marked by a check symbol. \textbf{c,} In the robot--robot setting with two morphologically identical robots, per-frame cosine similarity of the proprioceptive state to the self mask (proprio.--self) against its similarity to the other robot's mask (proprio.--other), before and after training. \textbf{d,} Distributions of synchronous (proprioception and robot mask from the same frame) and asynchronous (from different frames) cosine similarities, before and after training. \textbf{e,} t-SNE projections~\cite{van2008visualizing} of proprioceptive-state embeddings and humanoid robot mask embeddings before and after training.}
    \label{fig:self_other_distinction}
\end{figure*}

We first ask whether a humanoid robot can identify its own body in a multi-agent scene using only proprioception and vision. Given the robot's proprioceptive state (proprio.) and a set of candidate body masks from the visual frame, the task is to select the one that corresponds to its own body (Fig.~\ref{fig:self_other_distinction}b). Neither identity labels nor kinematic models are provided. The only learning signal is the temporal co-occurrence of proprioceptive states and visual observations, collected while the robot moves in scenes that also contain another agent.

The key intuition is that this correspondence holds only within a frame. Within the current frame, a specific candidate mask corresponds to the robot, though we do not know which. In a frame captured at a different time, the candidate masks are unlikely to match the current proprioceptive state, because the robot will generally have moved to a different configuration. To exploit this asymmetry, two encoders embed the proprioceptive state and each candidate mask into a shared space. The per-mask similarities are then fused via attention into a single frame-level score. We treat the frame paired in time with the current state as the positive and all other frames in the batch as negatives, yielding a fully self-supervised contrastive signal that never requires labelling the correct mask. The full architecture and training protocol are described in the Methods.
We train and test the model independently in two settings: 25,667 real-world human-robot frames, and, to probe whether self-identification survives a morphologically matched distractor, 9,000 simulated two-humanoid frames in which the distractor is another humanoid robot.

We compare our model against zero-shot vision-language model (VLM) baselines and a direct fusion variant in both evaluation settings (Fig.~\ref{fig:self_other_distinction}a). The VLM baselines include two open-source models, LLaVA v1.6~\cite{llava2023} and InstructBLIP~\cite{dai2023instructblip}, as well as two closed-source frontier models, GPT-5.5 and Gemini 3.1 Pro Preview. Our model exceeds 99.5\% accuracy in both, showing that the proposed approach robustly identifies the robot's own body across simulated and real-world settings. These VLMs, given the same visual scene and joint angles, perform at or only marginally above chance, indicating that current VLMs cannot bind a specific configuration of joint angles to the corresponding visible body. The direct fusion variant, which averages per-mask similarity scores rather than fusing them through attention, also degrades markedly: averaging gives the proprioception-uncorrelated distractor mask equal weight in the frame-level score, whereas attention lets the candidate most consistent with proprioception dominate. This gap is most pronounced in the human-robot setting. Figure~\ref{fig:self_other_distinction}b provides qualitative examples: across different poses, the model consistently selects the robot mask rather than the distractor mask.

An alternative explanation in the human-robot setting is that the model might rely on appearance rather than proprioception, since the human distractor looks different from the robot. To rule this out, we study the problem from two perspectives. First, we turn to the robot--robot setting, where the two candidates are morphologically identical and appearance offers no cue at all. In addition to the accuracy of $>$99.5\% in this setting (Fig.~\ref{fig:self_other_distinction}a), we examine the underlying similarities directly, comparing the proprioceptive feature against the self mask and against the other robot's mask in each frame; after training, the proprioceptive feature stays consistently closer to the self mask than to the other robot's mask in the same frame (Fig.~\ref{fig:self_other_distinction}c), so the model still selects the right body when the two are visually indistinguishable. Second, we ask whether the proprioceptive state aligns specifically with the visual observation from the same moment, comparing the synchronous similarity $\cos(\bm{f}^{i}_{\text{state}}, \bm{f}^{i}_{\text{image},\text{robot}})$, between the proprioceptive feature $\bm{f}^{i}_{\text{state}}$ and the robot mask from the same frame, against the asynchronous similarity $\cos(\bm{f}^{i}_{\text{state}}, \bm{f}^{j}_{\text{image},\text{robot}})$, between the same feature and the robot mask from a different frame $j \neq i$ (Fig.~\ref{fig:self_other_distinction}d); after training, the synchronous similarity is sharply higher than the asynchronous similarity. The same temporal correspondence is also visible in the embedding space: t-SNE projections~\cite{van2008visualizing} (Fig.~\ref{fig:self_other_distinction}e) show that before training, proprioceptive-state embeddings and robot mask embeddings occupy separate regions and even same-frame pairs lie far apart, whereas after training the two modalities collapse onto the same region and pairs from the same frame map to nearby points. Together, these results suggest that the model binds the current proprioceptive state to the robot configuration at that moment, rather than to a generic robot silhouette.

\subsection{Self-other distinction enables self-modeling}

We next ask whether proprioceptive--visual correspondence can also support self-modeling. A person can imagine the position of their limbs even with their eyes closed; can a robot do the same? We formulate this as predicting a continuous three-dimensional (3D) occupancy field of the robot's body from its joint state.

\begin{figure*}[!ht]
    \centering
    \includegraphics[width=0.98\linewidth]{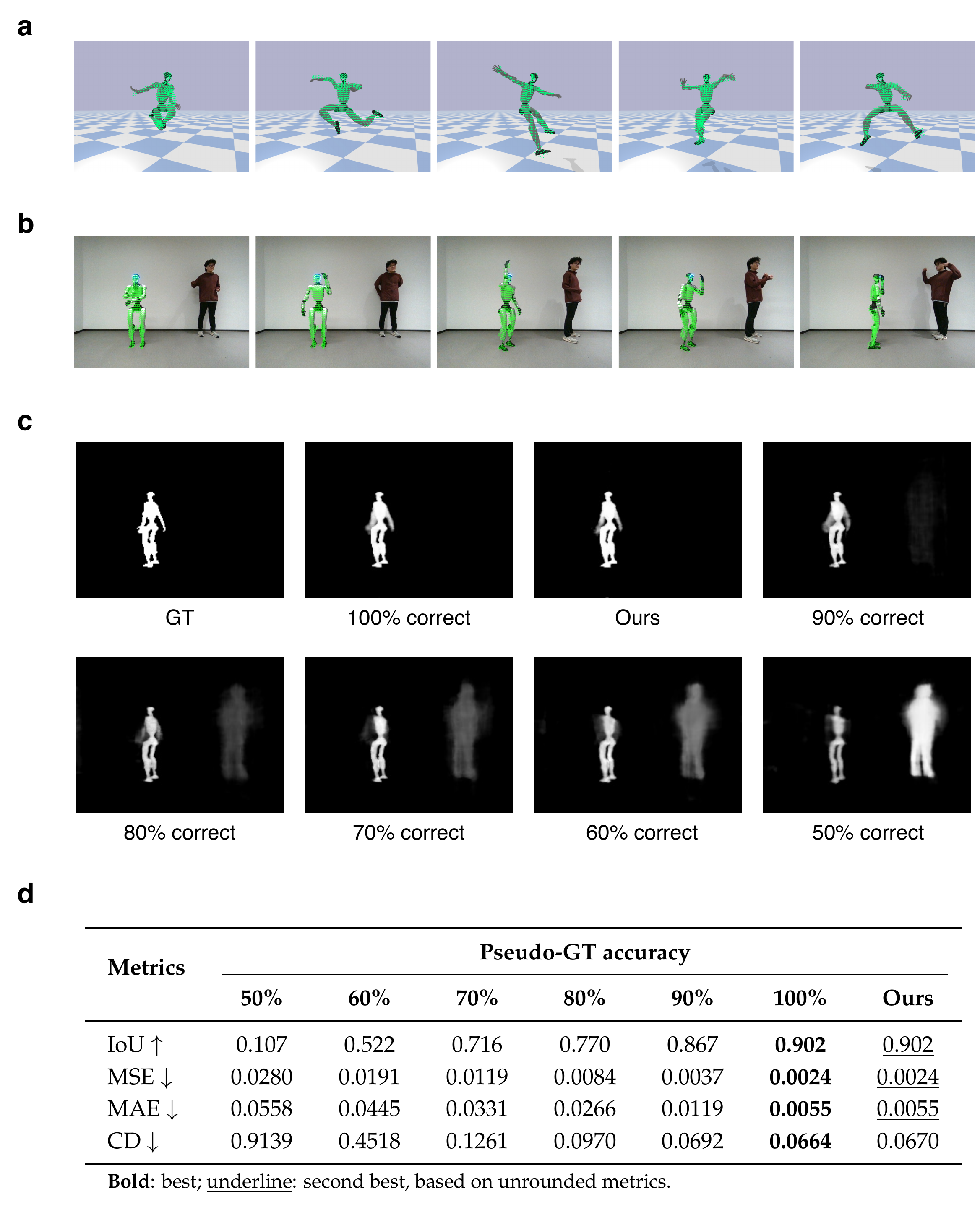}
    \caption{\textbf{Kinematics-free self-modeling from proprioception.} \textbf{a,} Simulated examples in which the learned self-model predicts robot body occupancy under different poses and viewpoints. \textbf{b,} Real-world examples in which the predicted occupancy is projected onto the robot while a human distractor remains excluded. \textbf{c,} Qualitative effect of the percentage of correct pseudo-GT masks on the learned self-model. At high pseudo-GT accuracy, the predicted robot body remains identifiable but begins to show ghost occupancy around the distractor; as accuracy falls further, the distractor response strengthens and the self-other boundary breaks down. \textbf{d,} Quantitative self-model fidelity as a function of the percentage of correct pseudo-GT masks, measured by IoU, MSE, MAE and Chamfer Distance. Performance with the pseudo-GT masks produced by self-other distinction ($>$99.5\% accuracy) closely matches the oracle trained on GT masks, whereas artificially degraded pseudo-GT masks lead to progressive deterioration.}
    \label{fig:self_reconstruction}
\end{figure*}

Concretely, we model the humanoid body with a pose-conditioned implicit field that maps a 3D query point to two scalar values: a body density predicting whether the point belongs to the robot, and a view-dependent visibility predicting, for a given camera ray, how much that point contributes to the observation. To make this formulation tractable for a 29-DoF humanoid, the proprioceptive state is encoded with a part-aware encoder that processes the torso and bilateral limbs separately, while sampled 3D points and viewing directions use sinusoidal positional encoding. We train the field from 2D self-mask supervision using a bounded volumetric mask renderer, which accumulates density and visibility only within the calibrated near--far range of the robot body. This adapts neural radiance field (NeRF)-style transmittance~\cite{2020NeRF,gao2022nerf} to binary silhouette supervision and avoids treating the far end of the ray as an unbounded absorber. Qualitative and quantitative comparisons with reconstruction variants and an adapted reconstruction baseline further show that both the part-aware architecture and the bounded visibility-aware renderer contribute to reconstruction fidelity across held-out poses (Supplementary Figs.~\ref{fig:supp_reconstruction_2d} and~\ref{fig:supp_reconstruction_results}).

The learned self-model produces coherent predictions in both simulation and the real world (Fig.~\ref{fig:self_reconstruction}a-b). To visualize the prediction, we threshold the learned density field to obtain a point cloud of the robot's body. This point cloud tracks the body across poses and orientations, and remains confined to the robot even when a human is present in the scene. The self-other boundary established in the previous section is therefore preserved in the 3D self-model.
Notably, the self-model is trained entirely from the pseudo-GT masks produced by self-other distinction. Here, ``Ours'' denotes training with the masks selected by the first-stage self-other distinction module. With self-other distinction at $>$99.5\% accuracy, the resulting fidelity of the self-model closely matches an oracle model trained on GT masks (Fig.~\ref{fig:self_reconstruction}c-d). These results show that it is possible to learn a faithful and stable self-model from imperfect, fully self-supervised supervision.

To probe how sensitive self-modeling is to the accuracy of self-other distinction, we artificially reduce the percentage of correct pseudo-GT masks by randomly flipping a fraction of self and other masks. Mild corruption still preserves the main body shape but introduces ghost occupancy around the distractor, whereas stronger corruption makes the predicted body occupancy increasingly diffuse and can cause the self-other boundary to collapse (Fig.~\ref{fig:self_reconstruction}c). The same trend appears quantitatively: all four metrics deteriorate as the percentage of correct pseudo-GT masks decreases (Fig.~\ref{fig:self_reconstruction}d). Self-other distinction is not merely a precursor to self-modeling, because its accuracy directly determines whether self-modeling can succeed.
The detailed design and full evaluation protocol are described in the Methods.

\subsection{From self-model to physical interaction}

Having shown that the robot can distinguish itself from others and learn a 3D representation of its own body, we next ask whether the knowledge of its own body can guide physical interaction with the real world. Each task can be cast as a spatial constraint on the robot's body, whether the goal is to reach a target, avoid an obstacle, or imitate a demonstrated pose. Because the self-model provides a differentiable mapping from joint configurations to 3D body occupancy, these constraints translate into gradients over joint angles, allowing the robot to solve for the configuration that satisfies the task. We test this on three tasks of increasing complexity: target reaching, collision-aware motion planning, and human-to-robot motion retargeting.

\begin{figure*}[!htb]
    \centering
    \includegraphics[width=0.98\linewidth]{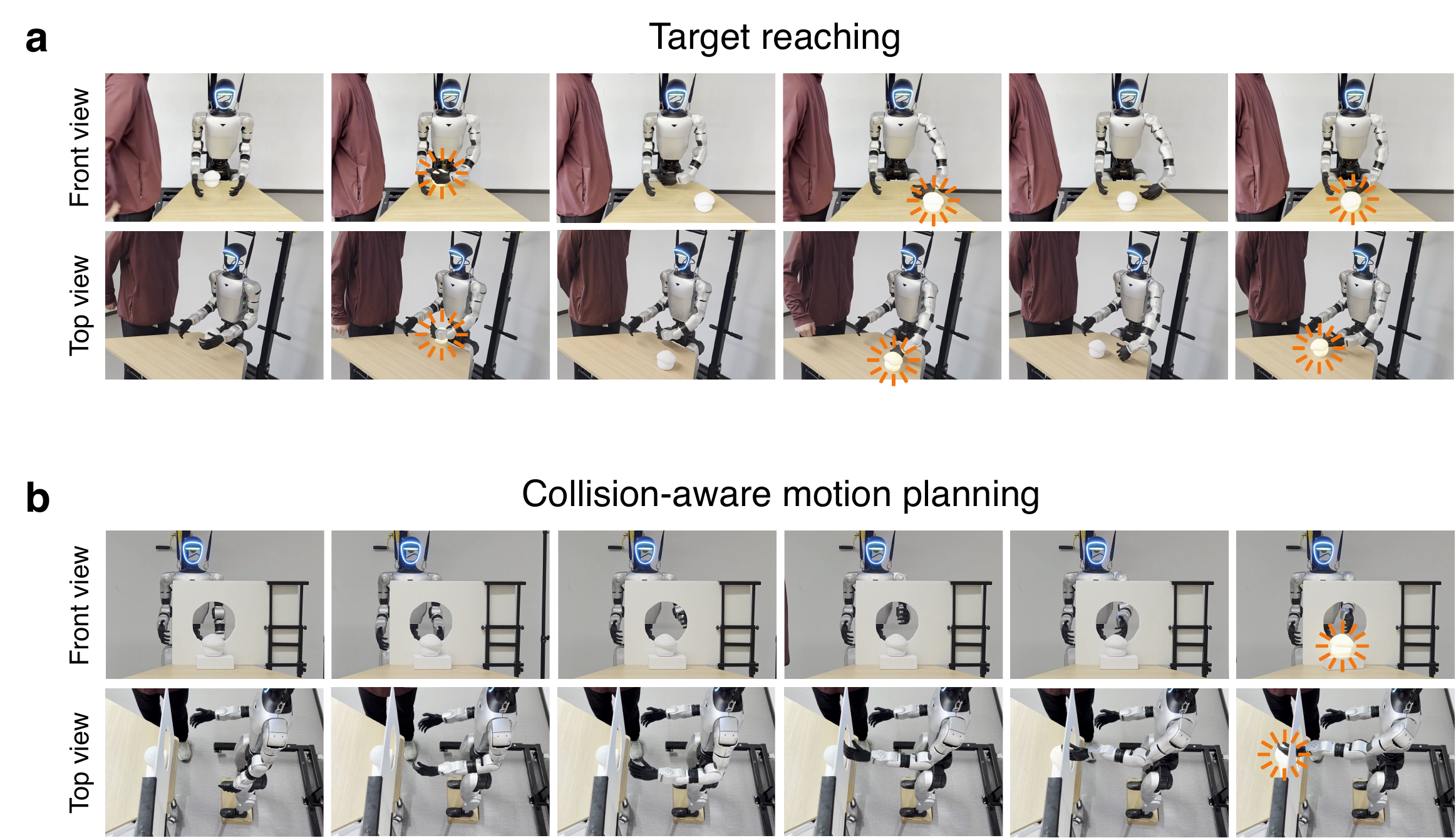}
    \caption{\textbf{Target reaching and collision-aware motion planning.} \textbf{a,} Target reaching. A human places a tabletop night-light at different positions, and the robot reaches and touches it to light it up, shown from the front and top views. \textbf{b,} Collision-aware motion planning through a board with a circular aperture. Starting with the left arm hanging down, the robot raises the arm, passes it through the opening, and reaches the night-light. A temporal sequence is shown from the front and top views.}
    \label{fig:downstream_reaching}
\end{figure*}

We begin with target reaching (Fig.~\ref{fig:downstream_reaching}a), the most direct test of the self-model as a control interface. Specifically, a human places a small night-light at different positions on a tabletop, and the robot optimizes its seven left-arm joint angles to bring the centre of a hand-specific self-model to the target. The robot hand is approximately 140\,mm long and 115\,mm wide, and physical tests showed that the night-light could be activated when the optimized hand centre approached within 100\,mm of the target. We therefore used 100\,mm as the task-level success threshold. We repeated the reaching test 50 times across different target placements. The mean best distance was 51.3\,mm, and 44 of 50 trials met this criterion, corresponding to an 88.0\% success rate. These results show that the self-model provides a differentiable geometric signal sufficient to guide physical reaching.

Reaching only constrains the position of the hand. A harder test is whether the self-model can also reason about the geometry of the entire arm when the environment imposes constraints along the way. We therefore turn to collision-aware motion planning (Fig.~\ref{fig:downstream_reaching}b), where the night-light is placed behind a $400\,\text{mm}\times600\,\text{mm}$ planar board with a circular aperture of radius $100\,\text{mm}$, so a direct gradient-descent trajectory toward the target would collide with the board. The seven left-arm joint angles are optimized as in the reaching task, but the learned full-body and hand occupancy models now serve as collision and goal queries inside a target-biased RRT planner. Across 14 constrained planning settings, 10 trajectories reached the target under the same 100\,mm task-level success threshold, corresponding to a 71.4\% success rate. Across all settings, the mean final distance was 89.3\,mm and the average path length was 155.4 waypoints. 

\begin{figure*}[!ht]
    \centering
    \includegraphics[width=0.98\linewidth]{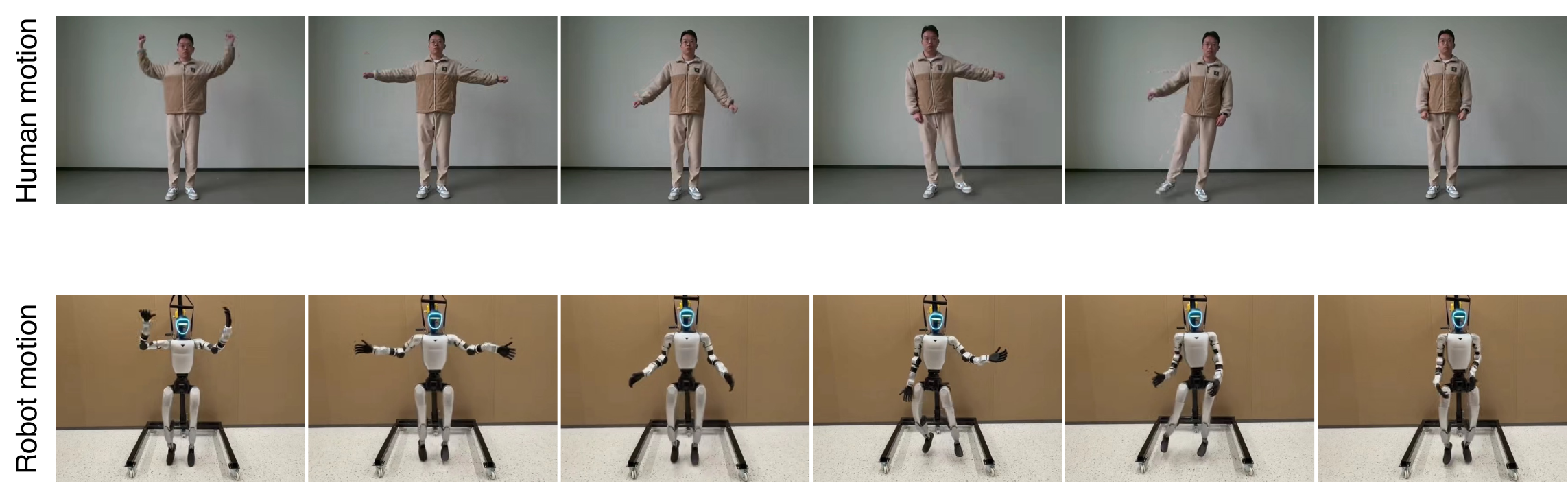}
    \caption{\textbf{Human-to-robot motion retargeting.} A human demonstration (upper row) is retargeted to a robot motion sequence (lower row) by optimizing the robot's joint configuration through the learned self-model.}
    \label{fig:motion_retargeting}
\end{figure*}

We further generalize from controlling a single arm to whole-body motion transfer through human-to-robot retargeting (Fig.~\ref{fig:motion_retargeting}). To support this, we extend the self-model with multiple part-specific output heads that share a common backbone, with one head for each body part of interest (e.g., hands and feet). More details can be found in Methods. Given a whole-body human demonstration, we extract 3D keypoints for these parts, map them to robot-compatible targets, and optimize the robot's full 29-DoF configuration so that each predicted body part reaches its target. Because the self-model is a geometric rather than physical representation, the robot is externally supported during retargeting (see Discussion). The resulting robot joint sequence reproduces the human whole-body motion. To quantify this, we measure the per-keypoint distance between the optimized robot body parts and their targets across 50 poses, obtaining a mean error of 36.1\,mm, corresponding to approximately 2.7\% of the robot body height (1320\,mm). This is achieved without any paired human-robot training data.

\section{Discussion}\label{sec:discussion}

We have shown that a humanoid robot can identify its own body in a multi-agent scene from proprioceptive--visual correspondence alone, and that this distinction in turn supplies the supervision for a kinematics-free 3D self-model. This builds on self-supervised self-modeling, previously studied in single-agent settings~\cite{hu2025teaching}. That prior work assumes the robot is the only body in the scene and demonstrates the idea on a 4-DoF robotic arm, so the question of which body to model does not arise. Our framework removes this assumption and addresses a fundamental question that emerges only in multi-agent settings: which body is the self? Selfhood, in this framework, is defined operationally: among the bodies visible in the scene, the self is the one whose visual configuration changes in agreement with the robot's internal state.

At first glance, self-other distinction and self-modeling appear to be mutually dependent. Yet proprioceptive--visual correspondence breaks this circularity by enabling the robot to recognize its own body without any prior body model, and the resulting distinction in turn bootstraps the supervision needed for self-modeling. The strength of this link is confirmed by the observation that self-model fidelity degrades progressively as self-other distinction accuracy is reduced, with severe degradation producing ghost occupancy and eventual loss of the self-other boundary. Notably, the same developmental ordering is observed in human infants, where perceptual self-other differentiation emerges well before detailed body representations~\cite{brownell2007}, suggesting that this bootstrapping sequence may reflect a general principle rather than a design choice.

From an engineering perspective, the robot's body region may be isolated through predefined geometric models or manual annotation, but these approaches presuppose precisely the prior knowledge of the robot's own body that self-modeling aims to acquire. The more fundamental scientific question is whether proprioceptive--visual correspondence alone, the same mechanism that biological agents rely on, is sufficient to ground both self-other distinction and self-modeling without any such priors. Our results suggest that it is. No explicit identity label or kinematic prior is needed; the temporal coupling between proprioceptive and visual streams is itself a sufficient self-supervised signal. This echoes neurocognitive accounts of body ownership, in which the sense of one's own body is constructed from the moment-to-moment congruence between proprioceptive and visual signals~\cite{kalckert2012}. The fact that even modern VLMs, despite their strong semantic capabilities, fail to recover this binding further supports the view that self-other distinction is not a semantic recognition problem but one that requires learning the cross-modal correspondence from experience.

Several aspects of the current study suggest natural extensions. First, candidate body masks are produced by an off-the-shelf segmentation and tracking model. In our experiments this front end was reliable enough that segmentation errors did not affect subsequent modeling, but in principle any upstream failure would propagate, and an end-to-end formulation is a natural way to remove this dependency. Second, our evaluation spans human distractors in the real world and morphologically identical humanoid distractors in simulation; scaling to denser scenes with more distractors is a natural next step that the same proprioceptive--visual mechanism should in principle support.
Third, the learned self-model is a geometric representation: it maps joint configurations to body occupancy but does not model physics. Consequently, whole-body retargeting is demonstrated with the robot externally supported, because standing balance requires physical reasoning about gravity and contact. Existing systems that achieve free-standing retargeting rely on manually provided kinematic and dynamic models; achieving the same from a learned geometric representation alone, without such priors, is a substantially harder problem that warrants dedicated investigation. Moreover, although we validate the framework on a humanoid robot, the underlying principle, proprioceptive--visual correspondence, is not specific to any particular morphology. The part-aware encoder groups joints by body region rather than relying on humanoid-specific structure, and could in principle be applied to other articulated robots.

Beyond these extensions, the capability demonstrated here occupies a specific position in the stack of social cognition. 
Higher-level social competences, such as imitation, coordination and theory of mind, all presuppose that the agent has already established a boundary between self and other, yet current work on these problems operates in settings where this boundary is predetermined. Our framework addresses this prerequisite directly: the learned self-other distinction, together with the self-model it supervises, provides a necessary foundation on which higher-level social competences could be built. More broadly, the building blocks of social cognition may also be acquired progressively from interaction, rather than engineered independently for each task.

Taken together, these results suggest that bodily self-representation need not be engineered or provided a priori, but can emerge from the cross-modal structure of sensorimotor experience alone. Extending this representation to denser social scenes, diverse robot morphologies, and physical reasoning would take a further step toward robots that, like biological agents, come to know their own bodies through experience and act alongside others in shared physical environments.

\clearpage

\section{Methods}\label{sec:method}

\begin{figure}[!htbp]
    \centering
    \includegraphics[width=0.98\linewidth]{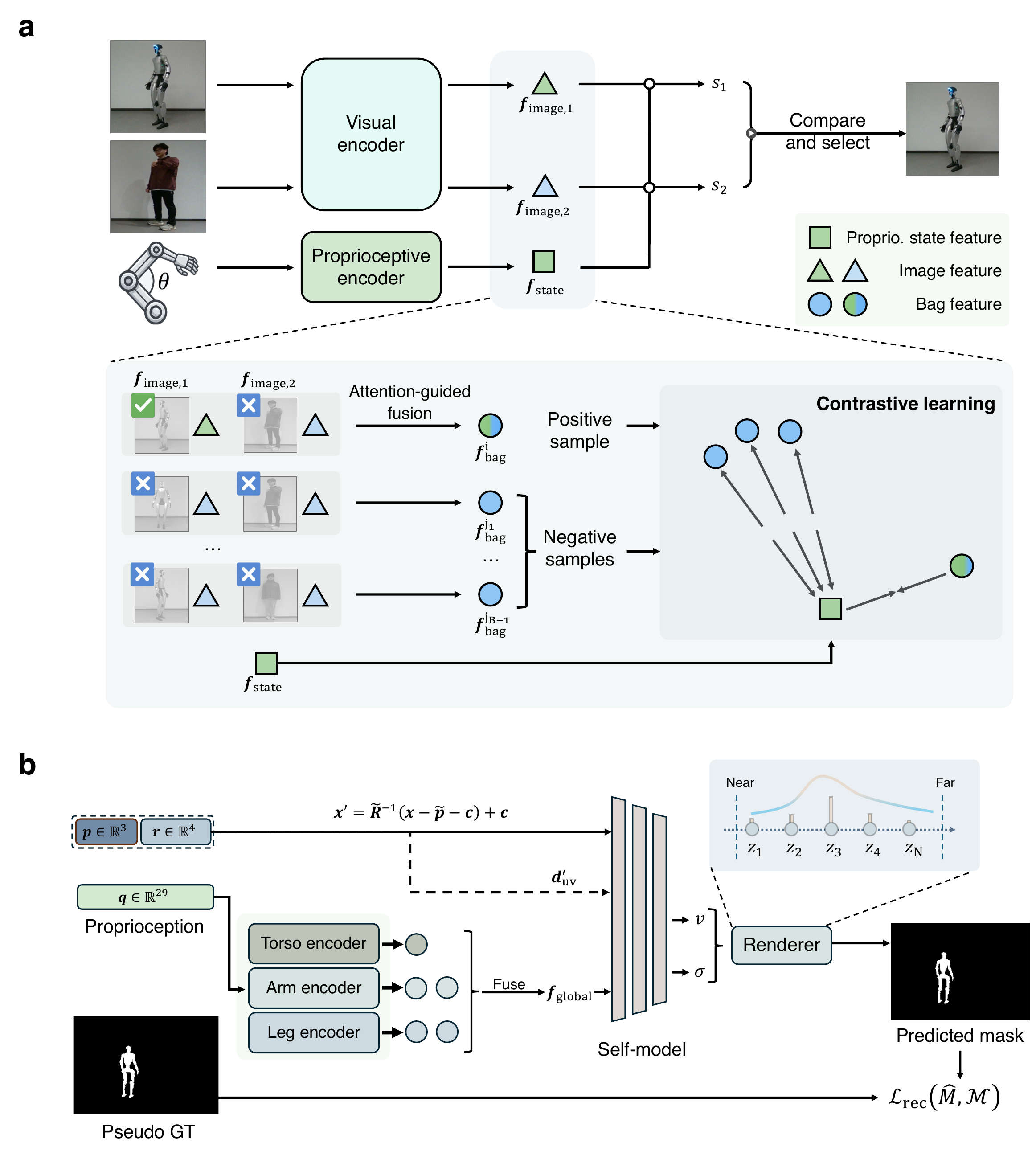}
    \caption{\textbf{Method overview.} \textbf{a,} Self-other distinction compares proprioceptive and visual embeddings, selects the self-mask, and trains the alignment with attention-guided contrastive learning. \textbf{b,} Kinematics-free self-modeling uses the selected self-mask to learn a pose-conditioned density and visibility field through bounded volumetric mask rendering.}
    \label{fig:method_pipeline}
\end{figure}

\subsection{Problem formulation and method overview}

We study kinematics-free self-modeling in a multi-agent visual scene. At time step $t$, our model takes as input an RGB image $\mathcal{I}_t$ captured by a single fixed-view external camera (Intel RealSense D455, RGB stream only) and the robot's proprioceptive state from odometry and joint sensing. The full proprioceptive state is
\begin{equation}
\mathcal{S}_t = [\bm{q}_t, \bm{r}_t, \bm{p}_t] \in \mathbb{R}^{36},
\end{equation}
where $\bm{q}_t \in \mathbb{R}^{29}$ denotes the 29-DoF joint configuration, $\bm{r}_t \in \mathbb{R}^{4}$ is the root orientation represented as a quaternion, and $\bm{p}_t \in \mathbb{R}^{3}$ is the root translation. The image may contain the robot body, humans, other robots, and the static scene. The robot is not given an identity label linking its proprioceptive state to any particular visual body, and it is not given a forward-kinematics model, link geometry, CAD model, or Unified Robot Description Format (URDF) file.

The method contains two modules (Fig.~\ref{fig:method_pipeline}). The self-other distinction module uses a 30-dimensional proprioceptive state, consisting of the 29 joint angles and the robot's global yaw, to select the visual candidate whose shape is consistent with the robot's current state. The self-modeling module then uses the selected self-mask as pseudo-GT supervision to train a 36-dimensional pose-conditioned implicit self-model. This model maps sampled 3D points, viewing directions, and the full robot state to body occupancy and visibility, yielding a differentiable representation that can be used for target reaching, collision-aware motion planning, and human-to-robot motion retargeting.

\subsection{Data collection and input preparation}

We collect real-world data with a Unitree G1 humanoid robot and a fixed Intel RealSense D455 RGB camera placed 1.93\,m in front of the robot at a height of 0.7872\,m. Each RGB frame is paired with the robot's proprioceptive state from odometry and joint sensing. The real-world human-robot dataset contains 25,667 frames with a human distractor in the scene. We additionally generate 9,000 two-humanoid frames in Isaac Gym~\cite{makoviychuk2021isaac} using the same front-facing camera distance and height, and visualize the simulated scenes in PyBullet~\cite{coumans2016pybullet}. Candidate masks are extracted from the RGB images before training. Training and validation frames are kept separate, and each training sample uses the same input format: a 30D proprioceptive state and two candidate masks.

\subsection{Self-other distinction from proprioception}
\label{sec:method_discrimination}

\textbf{Candidate construction.} Candidate masks are extracted from the image using class-agnostic segmentation with the Segment Anything Model (SAM)~3~\cite{carion2025sam}. Each training sample contains two candidate instances: one mask corresponding to the ego robot and one distractor mask, where the distractor can be either a human or another robot. Each single-instance mask has spatial size $174\times232$, and the two masks are placed into a combined $174\times464$ canvas. The model does not use semantic category labels during optimization, and the corresponding evaluation label is updated only for accuracy measurement.

We use separate pairing protocols for human and robot distractors. In the robot--robot simulated setting, simulated robot samples are divided into two disjoint pools and then randomly paired across pools, so that one robot instance serves as the ego candidate and the other serves as the distractor. In the real-world robot--human setting, each pair of robot and human masks comes from the same captured frame, preserving the real multi-agent scene observed by the camera. 

\textbf{Proprioceptive encoding.} The distinction stage excludes root translation to remove the influence of where the robot happens to stand in the scene. This design reflects the movement structure of humanoid robots and humans on the ground: global planar displacement should not determine identity, whereas global heading changes the projected silhouette. We therefore retain only the yaw angle from the quaternion $\bm{r}_t$ as the global proprioceptive component, while preserving the full-body joint configuration as the local proprioceptive component. The resulting state is
\begin{equation}
\boldsymbol{\theta}_t = [\bm{q}_t, \psi_t] \in \mathbb{R}^{30},
\end{equation}
where $\psi_t$ is the yaw extracted from $\bm{r}_t$. We encode the joint configuration and yaw together with a proprioceptive state encoder, producing an L2-normalized feature
$\bm{f}_{\text{state}}$.

\textbf{Visual instance encoding.} The two candidate masks are cropped from their respective half-canvas regions and passed independently through an instance encoder. We apply position normalization and scale normalization before visual encoding, reducing the influence of where an agent appears in the image and how large it appears under different camera distances or viewpoints. The normalized masks are then processed by a shared image encoder, implemented as a custom CNN encoder for mask regions followed by a projection head, to obtain L2-normalized instance features $\bm{f}_{\text{image},1}$ and $\bm{f}_{\text{image},2}$. Similarities are computed by dot product,
\begin{equation}
s_k = \bm{f}_{\text{state}}^{\top} \bm{f}_{\text{image},k}, \quad k \in \{1,2\}.
\label{eq:cosine_similarity}
\end{equation}

The predicted self instance is the candidate with the larger similarity. Evaluation accuracy is computed as the proportion of frames in which the argmax similarity selects the robot mask.

\subsection{Two-level self-supervised optimization}
\label{sec:two-level}

The main challenge in self-other distinction is that the correct instance is unknown during learning. In each frame, the robot knows its own proprioceptive state, but it does not know which candidate mask corresponds to that state. Conversely, the set of candidates from the same frame is guaranteed to contain the self instance in our training construction. We use this weak structure to form a two-level self-supervised objective. The inner level treats the candidate masks in a frame as a multiple-instance bag and softly selects the instance most compatible with the robot state~\cite{ITW:2018}. The outer level supplies the learning signal by contrasting the selected same-frame bag against bags from other frames, whose proprioceptive states are unlikely to match the anchor state when sampled from sufficiently different moments in the motion sequence.

\textbf{Inner level: latent self selection within a frame.} For sample $i$, the model first compares the proprioceptive feature $\bm{f}^{i}_{\text{state}}$ with each candidate visual feature $\bm{f}^{i}_{\text{image},k}$, obtaining feature similarities $s^i_k$ with Equation~\ref{eq:cosine_similarity}.
These similarities are converted into attention weights via
\begin{equation}
\alpha^{i}_{k} =
\frac{\exp(s^{i}_{k}/\tau_a)}
{\sum_{\ell=1}^{2}\exp(s^{i}_{\ell}/\tau_a)},
\end{equation}
where $\tau_a$ controls how sharply the model selects among the candidates. The frame-level visual representation is then formed as a weighted bag feature,
\begin{equation}
\bm{f}_{\text{bag}}^{i} = \sum_{k=1}^{2} \alpha^{i}_{k} \bm{f}^{i}_{\text{image},k}.
\end{equation}
This operation makes the unknown self-mask a latent variable rather than a supervised label. If a candidate consistently carries visual structure that covaries with the robot's joint state and heading, it can receive increasing attention across training. If a candidate is a distractor, its shape changes independently of the robot's proprioception and therefore cannot support stable cross-modal alignment.

\textbf{Outer level: frame-wise contrastive alignment.} The outer level turns this latent selection into a trainable objective. For an anchor state $\boldsymbol{\theta}_i$, the positive visual representation is the bag feature from the same frame, $\bm{f}_{\text{bag}}^{i}$. Bag features from other frames in the batch provide negatives because they are usually associated with different robot states; repeated or near-static poses can create occasional false negatives, but these cases are expected to be rare under diverse motion sampling. This temporal contrast is the key source of supervision: the model is rewarded when the bag selected from the current frame is closer to the current proprioceptive state than most bags selected from other frames. With batch size $B$, the InfoNCE~\cite{oord2018representation} objective is
\begin{equation}
\mathcal{L}_{\mathrm{InfoNCE}} =
-\frac{1}{B}\sum_{i=1}^{B}
\log
\frac{
\exp((\bm{f}^{i}_{\text{state}})^{\top}\bm{f}_{\text{bag}}^{i}/\tau_l)
}{
\sum_{j=1}^{B}
\exp((\bm{f}^{i}_{\text{state}})^{\top}\bm{f}_{\text{bag}}^{j}/\tau_l)
}.
\end{equation}
The denominator includes the positive same-frame bag and all other bags in the batch as negatives. This objective couples the two levels. The contrastive term cannot be minimized by selecting an arbitrary candidate within the frame, because the resulting bag feature must also be predictive of the robot's proprioceptive state relative to other frames. The attention mechanism therefore receives an indirect but consistent signal: selecting the robot mask improves same-frame proprioceptive--visual alignment, whereas selecting a human distractor weakens the positive pair and makes the frame harder to distinguish from negatives. In this way, instance-level self labels emerge from frame-level proprioceptive--visual correspondence.

For all self-other distinction experiments, we use $\tau_a=0.003$, $\tau_l=0.01$, batch size $B=32$, learning rate $0.001$, 100 training epochs and weight decay $0.01$. We use AdamW optimization~\cite{loshchilov2017decoupled} with gradient clipping for the self-other distinction module. Ablation experiments for self-other distinction are reported in Supplementary Table~\ref{tab:supp_ablation_results}.

\subsection{Kinematics-free implicit self-modeling}

After self-other distinction, the selected robot masks serve as pseudo-GT supervision for learning the robot's self-model. This second stage uses the full 36-dimensional state $\mathcal{S}_t$ because self-modeling requires both local articulation and root pose. The target supervision is a binary or near-binary foreground mask $\mathcal{M}_t \in \{0,1\}^{H\times W}$, obtained from the self-mask selected by the self-other distinction module. The self-modeling stage therefore learns from the visual instance that the robot has identified as its own, preserving the causal chain from self-other distinction to self-modeling.

\textbf{Pose normalization.} To reduce global drift in real robot data, the first frame of each sequence is used as a local reference. Let $\bm{R}_0$ and $\bm{p}_0$ denote the root rotation and translation in the first frame. We transform later translations and rotations into this local coordinate system:
\begin{equation}
\tilde{\bm{p}}_t = \bm{R}_0^{-1}(\bm{p}_t-\bm{p}_0),
\quad
\tilde{\bm{R}}_t = \bm{R}_0^{-1}\bm{R}_t.
\end{equation}
The relative rotation is converted back to a quaternion $\tilde{\bm{r}}_t$. As a sequence-level preprocessing step, and because $\bm{r}$ and $-\bm{r}$ represent the same rotation, we enforce temporal continuity by flipping the quaternion sign whenever its dot product with the previous quaternion in the same sequence is negative. The reconstruction state is then
\begin{equation}
\tilde{\mathcal{S}}_t=[\bm{q}_t,\tilde{\bm{r}}_t,\tilde{\bm{p}}_t]\in\mathbb{R}^{36}.
\end{equation}

\textbf{Ray sampling and pose-conditioned coordinates.} Camera rays are generated from calibrated intrinsics. For pixel $(u,v)$ with intrinsics $(f_x,f_y,c_x,c_y)$, the camera-frame direction is
\begin{equation}
\bm{d}_{uv}=
\left[
\frac{u-c_x}{f_x},
-\frac{v-c_y}{f_y},
-1
\right]^{\top}.
\end{equation}
Points are sampled along each ray between near and far planes, with $\text{near}=d_{\text{cam}}-n_f$ and $\text{far}=d_{\text{cam}}+n_f$. For depth sample $z_k$, the point is $\bm{x}_k=\bm{o}+z_k\bm{d}_{uv}$.

The sampled point is then transformed into a robot-centred canonical space using the relative root pose:
\begin{equation}
\bm{x}'_k = \tilde{\bm{R}}_t^{-1}(\bm{x}_k-\tilde{\bm{p}}_t-\bm{c})+\bm{c},
\quad
\bm{d}'_{uv} = \frac{\tilde{\bm{R}}_t^{-1}\bm{d}_{uv}}
{\|\tilde{\bm{R}}_t^{-1}\bm{d}_{uv}\|_2},
\end{equation}
where $\bm{c}=[0,-0.457,0.793]^{\top}$ is the robot-centred offset used to align the canonical sampling volume. The vertical component corresponds to the root-joint height of the standing robot, and the lateral component accounts for the robot standing 0.457\,m to the left of the scene centre in our camera view. This transformation lets the network model articulated body shape while factoring out global motion.

\textbf{Visibility-based implicit self-model.} The reconstruction network learns an implicit function
\begin{equation}
F_{\phi}(\bm{x}'_k,\bm{d}'_{uv},\tilde{\mathcal{S}}_t)
\mapsto (v_k,\sigma_k),
\end{equation}
where $\sigma_k\ge 0$ is volume density and $v_k\in[0,1]$ is view-dependent visibility. The input dimension is 42, consisting of a 3D point, the 36D robot state, and a 3D viewing direction. Spatial coordinates, pose groups, and viewing direction are encoded with sinusoidal positional encoding. The part-aware pose encoder partitions the 29 joint angles into torso, left leg, right leg, left arm, and right arm groups, processes each group with a part-specific MLP, and fuses the resulting features into a global posture representation. This part-aware pose encoding provides a structured inductive bias for the high-dimensional humanoid state without imposing an explicit kinematic model. Bilateral limbs use shared structure in the encoder. The density branch predicts body geometry, whereas the visibility branch models view-dependent mask contribution.

\textbf{Bounded volumetric mask rendering and objective.} For consecutive samples along a ray, let
\begin{equation}
\delta_k =
\begin{cases}
z_{k+1}-z_k, & k<N.\\
\text{far}-z_N, & k=N.
\end{cases}
\end{equation}
This bounded volumetric mask renderer follows NeRF-style alpha compositing through the transmittance term $T_k$, but adapts the terminal interval to binary silhouette supervision. Standard RGB NeRF~\cite{2020NeRF} implementations often assign an effectively infinite distance to the final sample so that the ray can terminate beyond the sampled points. In our setting, the reconstruction volume is finite and calibrated around the robot body. We therefore set the last interval to $\text{far}-z_N$ rather than an unbounded value. This bounded terminal interval restricts the model to explaining occupancy within the near--far robot volume and avoids encouraging spurious opacity at the far end of background rays. Additional rendering comparisons are provided in Supplementary Figs.~\ref{fig:supp_reconstruction_2d} and~\ref{fig:supp_reconstruction_results}.

Density is converted to alpha by
\begin{equation}
\alpha_k = 1-\exp(-\sigma_k\delta_k),
\end{equation}
and transmittance is
\begin{equation}
T_k = \prod_{j<k}(1-\alpha_j).
\end{equation}
The predicted foreground value for ray $\boldsymbol{\rho}$ is
\begin{equation}
\hat{M}(\boldsymbol{\rho}) = \sum_{k=1}^{N} T_k\alpha_k v_k.
\end{equation}
We can also render the foreground mask without the visibility term, using $\sum_k T_k\alpha_k$, which still provides a strong reconstruction signal from density alone. The main reconstruction loss is
\begin{equation}
\mathcal{L}_{\text{rec}} =
\frac{1}{|\mathcal{R}|}
\sum_{\boldsymbol{\rho}\in\mathcal{R}}
\left(\hat{M}(\boldsymbol{\rho})-\mathcal{M}(\boldsymbol{\rho})\right)^2.
\end{equation}
Here, $\mathcal{R}$ denotes the set of sampled camera rays used for mask supervision, and $\boldsymbol{\rho}$ denotes an individual ray.
The self-model is trained mainly with this MSE objective, with a weak ray-wise density smoothness term used in real-world training to stabilize the learned field. Models are optimized with Adam~\cite{kingma2014adam} and ReduceLROnPlateau scheduling. For the comparison with FFKSM~\cite{hu2025teaching}, we adapt the original formulation from its 4-DoF robotic-arm setting to our 29-DoF humanoid setting by enlarging the pose input dimension and scaling the network depth and feature width accordingly, and train the adapted model on the same selected self-masks, camera calibration, data split and evaluation metrics as our own model.

\textbf{Pseudo-GT accuracy ablation.} To evaluate how errors in first-stage self-other distinction affect second-stage self-modeling, we construct controlled pseudo-GT mask sets with different fractions of correct self-masks. The 100\% condition uses oracle robot masks as supervision. For lower-accuracy conditions, we randomly replace a prescribed fraction of robot masks with the corresponding distractor masks, yielding pseudo-GT accuracies of 90\%, 80\%, 70\%, 60\% and 50\%. The Ours condition uses the masks selected by the trained self-other distinction module. All conditions use the same self-model architecture, training protocol, camera calibration and evaluation split. Reconstruction quality is evaluated on 6,013 test samples using IoU at an occupancy threshold of 0.5, MSE and MAE against GT masks, and Chamfer Distance against GT 3D point clouds.

\subsection{Downstream spatial reasoning with the learned self-model}
\label{sec:method_downstream}

The learned implicit self-model provides a differentiable mapping from proprioceptive state to body occupancy. We use this property to evaluate whether the representation can support human-related downstream tasks, rather than only reconstructing the robot's appearance. We consider three applications in human-robot collaborative settings: target reaching toward objects placed by a human, collision-aware motion planning through a constrained aperture, and human-to-robot motion retargeting.

\textbf{Left-hand self-model for target reaching.} For object-reaching tasks, the relevant occupied volume is the robot hand rather than the entire body. We therefore train a dedicated left-hand self-model that predicts the occupancy density of the left hand conditioned on the full-body proprioceptive state. Because the hand occupies only a small part of the full humanoid volume, we represent its 3D position by the density-weighted centre of the learned hand occupancy. Given query points $\{\bm{x}_i\}_{i=1}^{N}$ and predicted hand density $\sigma_{\text{hand}}(\bm{x}_i,\mathcal{S})$, the hand position is
\begin{equation}
\hat{\bm{c}}_{\text{hand}}(\mathcal{S}) =
\frac{\sum_{i=1}^{N} \mathrm{ReLU}\left(\sigma_{\text{hand}}(\bm{x}_i,\mathcal{S})\right)\bm{x}_i}
{\sum_{i=1}^{N} \mathrm{ReLU}\left(\sigma_{\text{hand}}(\bm{x}_i,\mathcal{S})\right) + \epsilon}.
\label{eq:hand_com}
\end{equation}
Given a target point $\bm{p}^{*} \in \mathbb{R}^{3}$ on the tabletop, the robot optimizes the seven left-arm joints while keeping the remaining state components fixed, yielding a full-body state $\mathcal{S}(\bm{q}_{\text{arm}})$ that minimizes
\begin{equation}
\mathcal{L}_{\text{target}} =
\left\|
\hat{\bm{c}}_{\text{hand}}(\mathcal{S}(\bm{q}_{\text{arm}})) - \bm{p}^{*}
\right\|_{2}.
\label{eq:target_reaching}
\end{equation}
Gradients are back-propagated through the learned hand self-model to update the joint angles. This procedure is a neural inverse-kinematics objective defined directly on the learned self-model, without using a predefined forward-kinematics chain or URDF-based controller.

\textbf{Collision-aware joint-space planning.} Target reaching becomes more difficult when the hand must move through a constrained environment. For the aperture-reaching task, the workspace contains a $400\,\text{mm}\times600\,\text{mm}$ planar board with a circular aperture of radius $100\,\text{mm}$, placed between the robot and the target so that a direct gradient-descent trajectory toward the target would collide with the board. We use learned occupancy models inside a target-biased rapidly exploring random tree (RRT)~\cite{LaValle1998RapidlyexploringRT}. Each node in the tree is a joint vector. At each planning step, the planner samples a candidate configuration, extends the nearest valid node toward that sample, and queries both the full-body self-model and the left-hand self-model at the candidate state. The full-body model checks whether any occupied body region intersects the planar board outside the circular aperture, while the left-hand model provides an additional hand-level collision query. The left-hand model is also used to test whether the hand has reached the target region. Candidate states that violate the collision checks are rejected while accepted states are added to the tree until the left hand reaches the target. This combines global search in joint space with collision and goal information supplied by the learned self-models.

\textbf{Human-to-robot motion retargeting.} For human-to-robot motion retargeting, we adapt the self-model from single-hand reaching to whole-body retargeting. The full pipeline is illustrated in Supplementary Fig.~\ref{fig:supp_motion_retargeting}. A monocular video of a human demonstration is first processed with a human pose estimator~\cite{yin2025smplest,araujo2025retargeting} to obtain a sparse sequence of 3D body keypoints. We use 11 keypoints: the pelvis, left and right shoulders, elbows, hands, knees, and feet (ankle joints). These keypoints are transformed from the camera coordinate frame into the robot coordinate frame. Because human and robot limb lengths differ, we do not directly use the raw human keypoint positions as targets. Instead, we preserve the direction of each human body segment and rescale the corresponding segment to the humanoid robot's body dimensions, producing robot-compatible spatial targets for each key body part.

The retargeted spatial targets do not directly specify robot joint angles. We therefore solve a neural inverse-kinematics problem using learned body-part occupancy models. Given a candidate 29-DoF joint vector $\bm{q}$, the model predicts a density field for each body part. We summarize each predicted body part by its density-weighted centre, using the same form as Equation~\ref{eq:hand_com}. For a set of retargeted target keypoints $\{\bm{c}^{*}_k\}_{k=1}^{K}$, the optimized keyframe pose minimizes
\begin{equation}
\mathcal{L}_{\text{retarget}}(\bm{q}) =
\frac{1}{K}\sum_{k=1}^{K}
\left\|
\hat{\bm{c}}_k(\bm{q})-\bm{c}^{*}_k
\right\|_2.
\label{eq:retarget}
\end{equation}
Gradients are back-propagated through the learned occupancy models to update the 29 robot joint angles, which are clamped to feasible ranges after each step. We optimize a sparse set of keyframes rather than every video frame, initializing each keyframe optimization from the same nominal robot pose.

\bibliography{sn-bibliography}
\bibliographystyle{colm2026_conference}

\ifincludesupplement
\clearpage

\clearpage
\section*{Supplementary Information}

\setcounter{figure}{0}
\setcounter{table}{0}
\renewcommand{\figurename}{Supplementary Fig.}
\renewcommand{\tablename}{Supplementary Table}
\renewcommand{\theHfigure}{supp.\arabic{figure}}
\renewcommand{\theHtable}{supp.\arabic{table}}

\begin{figure}[!htbp]
    \centering
    \includegraphics[width=0.98\linewidth]{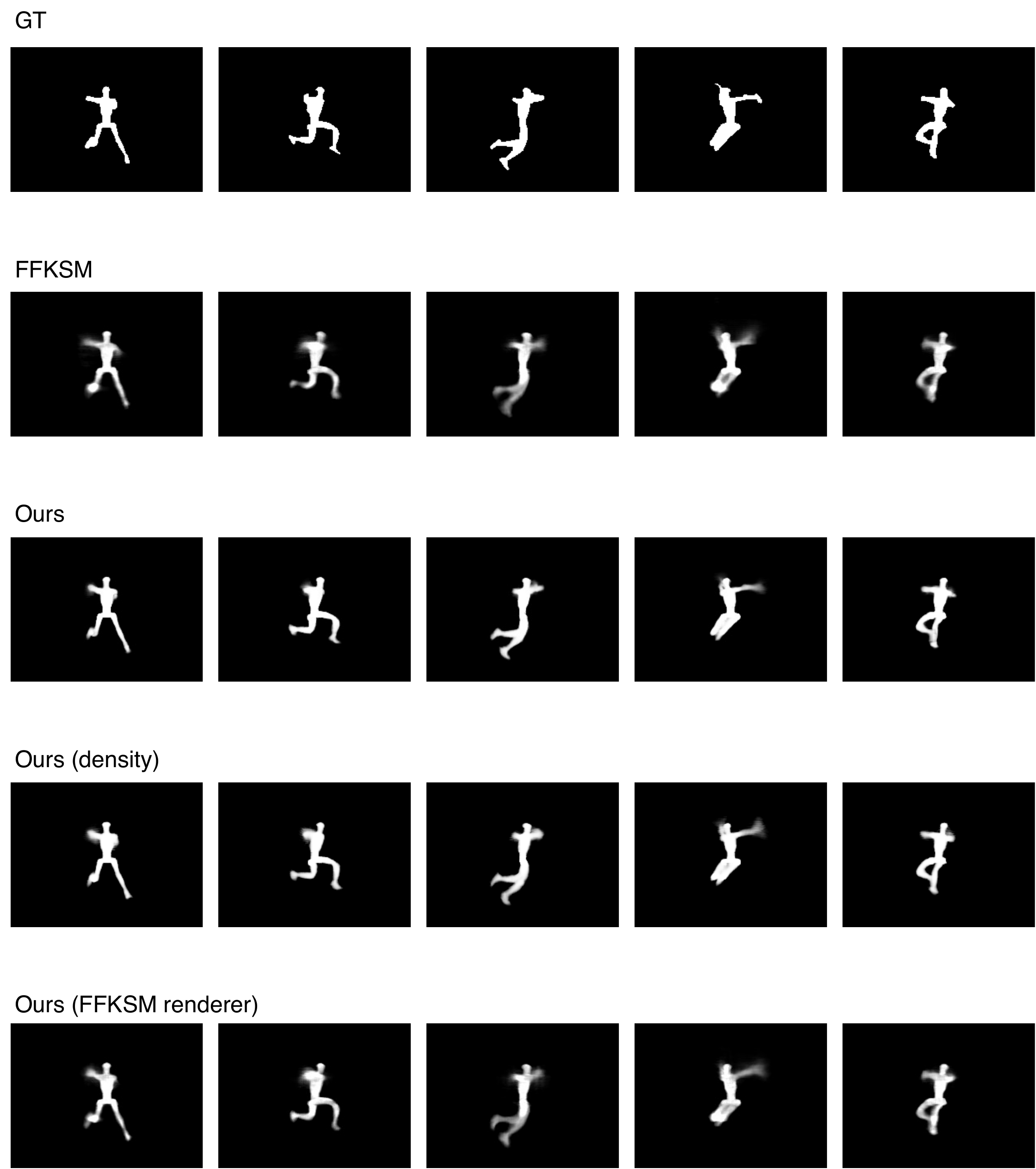}
    \caption{\textbf{Qualitative comparison of 2D reconstruction variants.}
    Rows show, from top to bottom, the GT mask, a scaled FFKSM baseline, our full method, our method without the visibility branch (Ours (density)), and our architecture combined with the FFKSM renderer (Ours (FFKSM renderer)). The comparison shows that the full model most closely matches the target silhouette, while the density-only variant remains competitive but loses view-dependent refinement. The FFKSM baseline and the variant that keeps our architecture but replaces the renderer with the FFKSM renderer indicate that both the part-aware reconstruction network and the bounded visibility-aware rendering formulation contribute to the final reconstruction quality.}
    \label{fig:supp_reconstruction_2d}
\end{figure}

\clearpage

\begin{figure}[!htbp]
    \centering
    \includegraphics[width=0.98\linewidth]{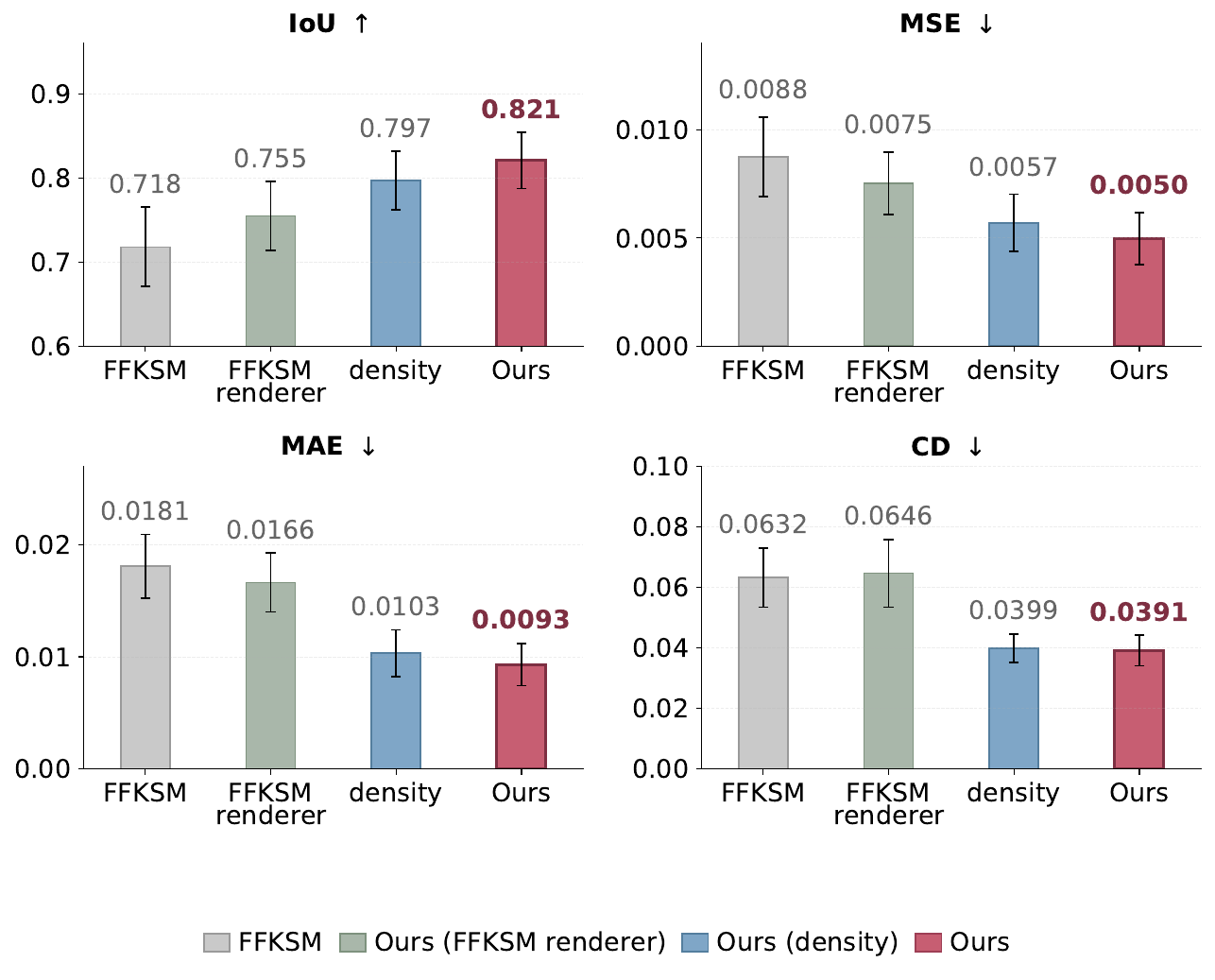}
    \caption{\textbf{Quantitative comparison of 2D reconstruction variants.}
    Quantitative evaluation of the same reconstruction variants shown in Supplementary Fig.~\ref{fig:supp_reconstruction_2d}: the scaled FFKSM baseline, our full method, our method without the visibility branch (Ours (density)), and our architecture combined with the FFKSM renderer (Ours (FFKSM renderer)). Reconstruction fidelity is measured by IoU, MSE, MAE and Chamfer Distance (CD), with higher IoU and lower MSE, MAE and CD indicating better performance. Bars show mean values, and error bars denote standard deviation across held-out evaluation samples. The full model achieves the best overall reconstruction accuracy, supporting the contribution of both the part-aware reconstruction architecture and the bounded visibility-aware rendering formulation.}
    \label{fig:supp_reconstruction_results}
\end{figure}

\clearpage

\begin{figure}[!htbp]
    \centering
    \includegraphics[width=0.98\linewidth]{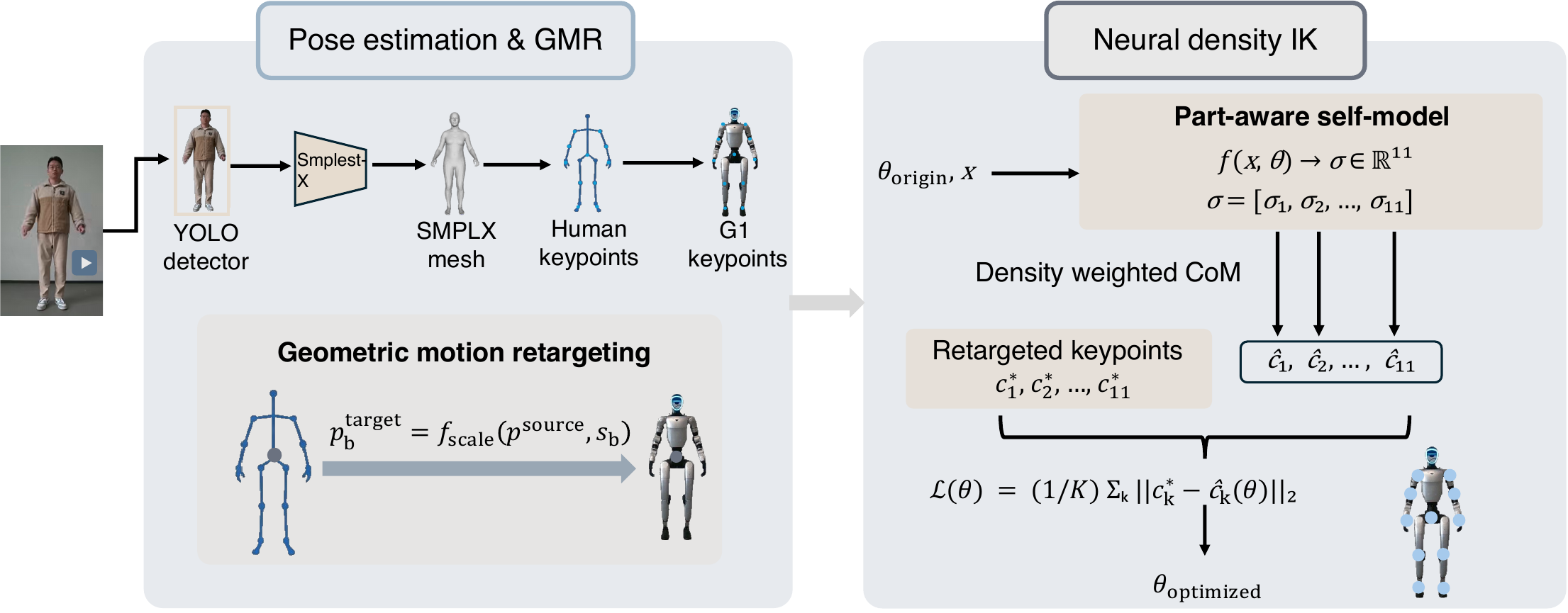}
    \caption{\textbf{Human-to-robot motion retargeting pipeline.}
    A monocular human demonstration is first processed by a human pose estimator to obtain a sparse sequence of 3D body keypoints. The 11 keypoints used for retargeting include the pelvis, left and right shoulders, elbows, hands, knees and feet; these points are transformed into the robot coordinate frame and mapped to robot-compatible spatial targets by preserving human segment directions while rescaling segment lengths to the humanoid robot body. The retargeted targets are then converted into robot joint configurations through neural inverse kinematics on the learned self-model. Given a candidate 29-DoF joint vector, the part-aware self-model predicts density fields for target body parts, whose density-weighted centres serve as differentiable robot keypoints. Gradients are back-propagated through the learned occupancy models to minimize the distance between predicted robot keypoints and retargeted targets, producing a whole-body robot motion sequence without paired human-robot demonstrations or a manually specified human-robot joint correspondence. In the diagram, $\bm{p}^{\mathrm{source}}$ denotes human keypoints, $\bm{p}^{\mathrm{target}}$ denotes robot-scaled target keypoints, $\theta$ denotes the robot joint vector, $\sigma_k$ is the predicted density for body part $k$, $\hat{\bm{c}}_k$ is the density-weighted robot keypoint predicted by the self-model, and $\bm{c}^{*}_k$ is the corresponding retargeted target.}
    \label{fig:supp_motion_retargeting}
\end{figure}

\clearpage

\begin{table}[!htbp]
\centering
\caption{\textbf{Comprehensive ablation study for self-other distinction.}
Validation accuracy is reported for four design choices in the self-other distinction module. Each block varies one factor while keeping the remaining settings fixed, testing the effect of representation dimension, scale normalization, attention temperature, and negative sampling strategy. Higher accuracy indicates better performance.}
\label{tab:supp_ablation_results}
\footnotesize
\setlength{\tabcolsep}{5pt}
\renewcommand{\arraystretch}{1.08}
\begin{tabular}{@{}ccccc@{}}
\toprule
\textbf{Dim} & \textbf{Scale norm} & \textbf{Attn temp} & \textbf{Sampling} & \textbf{Val acc.} \\
\midrule
\multicolumn{5}{l}{\textit{Feature dimension ablation}} \\
8 & ON & 0.003 & All & 0.7538 \\
16 & ON & 0.003 & All & \textbf{0.9969} \\
32 & ON & 0.003 & All & 0.9340 \\
64 & ON & 0.003 & All & 0.9860 \\
\midrule
\multicolumn{5}{l}{\textit{Scale normalization ablation}} \\
16 & ON & 0.003 & All & \textbf{0.9969} \\
16 & OFF & 0.003 & All & 0.8974 \\
\midrule
\multicolumn{5}{l}{\textit{Attention temperature ablation}} \\
16 & ON & 0.003 & All & \textbf{0.9969} \\
16 & ON & 0.03 & All & 0.8582 \\
16 & ON & 0.3 & All & 0.6884 \\
\midrule
\multicolumn{5}{l}{\textit{Negative sampling ablation}} \\
16 & ON & 0.003 & All & \textbf{0.9969} \\
16 & ON & 0.003 & Single & 0.8765 \\
\bottomrule
\end{tabular}
\end{table}

\clearpage

\fi

\end{document}